\documentclass{Interspeech}

\usepackage{hyperref}
\usepackage{graphicx}
\graphicspath{ {./images/} }   
\newcommand{\ignore}[1]{}

\usepackage[skip=0pt]{caption}

\usepackage{tablefootnote}

\usepackage{multirow}

\usepackage{epsfig}
\usepackage{caption}
\usepackage{subcaption}
\usepackage{algorithm}
\usepackage{algpseudocode}
\usepackage{amsmath}

\interspeechcameraready

\title{Understanding Dementia Speech Alignment with Diffusion-Based Image Generation}
\author[equalcontribution]{Mansi}{}\email{m.-24@imperial.ac.uk, a.lepipas20@imperial.ac.uk, d.woszczyk19@imperial.ac.uk, yiying.guan23@imperial.ac.uk, s.demetriou@imperial.ac.uk}
\author[equalcontribution]{Anastasios}{Lepipas }
\author[]{Dominika}{Woszczyk}
\author[]{Yiying}{Guan}
\author[]{Soteris}{Demetriou}
\affiliation[nocounter]{}{Imperial College London}{UK}

\usepackage{comment}

\begin{document}

\maketitle

\begin{abstract}

Text-to-image models generate highly realistic images based on natural language descriptions and millions of users use them to create and share images online. 
While it is expected that such models can align input text and generated image in the same latent space little has been done to understand whether this alignment is possible between pathological speech and generated images. In this work, we examine the ability of such models to align dementia-related speech information with the generated images and develop methods to explain this alignment. Surprisingly, we found that dementia detection is possible from generated images alone achieving $75\%$ accuracy on the ADReSS dataset. We then leverage explainability methods to show which parts of the language contribute to the detection.

\ignore{
Text--to--image models generate highly realistic images based on natural language prompts. Their accessibility has led to widespread adoption, with millions of users leveraging these models to create and share images online. In this work, we are the first to understand the risk of dementia--related information leakage from text--to--image diffusion models. Existing research on dementia inference has primarily focused on textual data. However, we demonstrate that image descriptions serve as a novel form to reveal this sensitive attribute. To illustrate this, we develop a new adversarial model that utilizes image embeddings\ignore{ to detect dementia--related leakage}, achieving $75\%$ accuracy on the ADReSS dataset. Also, we are the first to identify which are the salient points that contribute to the classification as well as it explores which information units are being leaked.

}
    
\end{abstract}
\section{Introduction}
\label{sec:introduction}

\vspace{5pt}\noindent\textbf{Problem Statement.} The rapid advancement of text-to-image (T2I) diffusion models, such as Stable Diffusion~\cite{Stable_Diffusion} has significantly expanded their applications, enabling highly realistic and contextually rich image generation from textual descriptions. By simply describing the desired image in natural language, users can leverage T2I models to produce visually accurate representations. However, as these models become increasingly integrated into speech and assistive technologies, critical privacy concerns emerge—-particularly in the context of vulnerable populations, such as individuals suffering from dementia.

This study investigates an unexplored privacy risk: the potential leakage of dementia-related characteristics through the output space of T2I diffusion models. Specifically, we explore whether images generated from speech-derived textual descriptions(from the ADReSS dataset) can inadvertently encode and reveal cognitive decline markers. Since dementia affects speech patterns like lexical choice, syntactic complexity, and fluency, diffusion models trained on text representations of speech may implicitly learn and propagate these features into their outputs. In our work, we focus our analysis on \emph{Information Units} (IU) which we define as the set of nouns and verbs which are present in the image, similar to \cite{zhu2023evaluating}; and \emph{discourse tokens} (pauses and filler words, e.g.: "um","uh", etc). Alarmingly, we show that diffusion models can inadvertently infer and externalize cognitive health indicators that can result in unauthorized profiling, discrimination, or stigmatization with generated images potentially carrying sensitive neurocognitive information and posing ethical, security, and privacy concerns.


\ignore{

}

\vspace{5pt}\noindent\textbf{Prior works.} 
The field of dementia classification from speech~\cite{luz2020alzheimer} has grown significantly, particularly since the $2020$ \textit{ADReSS} challenge. 
These models can be categorized into three main types: those using exclusively acoustic data~\cite{chlasta2021towards, mahajan2021acoustic}, those only using speech transcriptions~\cite{mahajan2021acoustic, millington2021analysis, balagopalan2020bert, guo2021crossing}, and hybrid models combining both~\cite{yuan2020disfluencies, zhu2022domain, haulcy2021classifying, martinc2021temporal}. 
More related to our dementia inference classifiers are models trained on transcriptions. Among those, BERT--based models have been shown to be highly effective yielding $81\%$--$83\%$ accuracy~\cite{balagopalan2020bert}.\ignore{ We replicate one of those models for our input attack.} More similar to our approach\ignore{ output dementia adversary model} are the models by Zhu et al.~\cite{zhu2021exploring} which combine 
information from the original ``Cookie Theft Picture'' with transcriptions to achieve between $80.63\%$ and $89.6\%$ accuracy on the \textit{ADReSS} dataset which matches the SOTA result ($89.6\%$) from Yuan et al's ERNIE-based text--audio approach~\cite{yuan2020disfluencies}. The main scope of our work is not to improve the dementia inference accuracy but to understand whether pathological speech information leakage is possible in image generation.

\ignore{Similarly with Zhu et al.~\cite{zhu2021exploring} and Yuan with other works we also utilize text and image information in our Input--Output attack. However, compared with Zhu et al.~\cite{zhu2021exploring} our $\mathcal{A}^{+}_{io}$ achieved better results without prior knowledge of the image the participants are trying to describe and instead utilize the output of T2I models.}

In addition, previous works have focused on analysing the input space features which lead to dementia classification. Model agnostic explainability techniques like SHAP~\cite{SHAP} and LIME~\cite{LIME} have been used to identify linguistic patterns~\cite{9769980} as well as key attributes that influence classification outcomes~\cite{Viswan2024, Alatrany2024}. Other works~\cite{Jahan2023, app14188287} leverage techniques like GradCAM and SHAP on multimodal data to identify the key aspects responsible for distinguishing dementia. To the best of our knowledge, there has been no work towards analysing the relationship of these characteristics across modalities in generative models.

\vspace{5pt}\noindent\textbf{Our Approach.} To investigate the potential leakage of dementia-related information through T2I diffusion models, we use a three-stage analysis framework which consists of speech-to-text conversion, text-to-image generation, and image-based inference analysis. This approach allows us to study the relationship between sensitive information from speech and generated images, and evaluate the extent to which diffusion models encode and propagate cognitive decline markers.
We leverage the transcription samples from the ADReSS dataset which resemble natural image descriptions as input prompts to a diffusion model to instruct it to generate images.


\vspace{3pt}\noindent Below we summarize the \textbf{main contributions} of this work:

\vspace{3pt}\noindent$\bullet$\textbf{ Novel Application Domain.} To the best of our knowledge, this is the first study to explore dementia-related speech leakage in T2I diffusion models. By bridging pathological speech analysis with generative AI security, we highlight a previously overlooked privacy dimension for a vulnerable population.


\vspace{3pt}\noindent$\bullet$\textbf{ New Dementia Inference Models.} We demonstrate that images generated from dementia-affected speech descriptions can encode implicit markers of neurocognitive decline posing a risk to individuals' privacy.  


\vspace{3pt}\noindent$\bullet$\textbf{ New Understanding.} We demonstrate that dementia can be leaked through T2I model outputs, with background details being the most discriminating features. Additionally, discourse tokens impact detection but are not the sole cause of leakage.


\ignore{We also reveal how and why the most influential features for classification tend to transform into background elements.}
 
\section{Threat Model}
\label{sec:dementia_detection}

\label{subsec:Threat_model}

Our threat model focuses on T2I diffusion models.
Such models leverage natural language descriptions
to guide the generation process.
Therefore, we hypothesize that they might
inadvertently encode linguistic markers of cognitive decline into their visual outputs.
This raises significant privacy concerns, as generated images may serve as unintended carriers of sensitive neurocognitive information, allowing adversaries to infer a speaker’s cognitive status without direct access to their speech or text. The ability of T2I models to propagate such information underscores the urgent need for privacy-preserving mechanisms in generative AI applications for healthcare. To examine our hypothesis we focus on dementia-related speech transcriptions. Note that real systems already exist which take speech as input, then convert it to text before passing it to the T2I model~\cite{ashutosh2025}.


We consider an adversary $\mathcal{A}$ with black--box access to T2I models, as it reflects a more realistic attack scenario compared to a white--box setting, which would require access to the model’s internal components. We define two adversary models with different levels of access to resources, text--based ($\mathcal{A}_{i}$) and image--based ($\mathcal{A}_{o}$) only. Considering the related work, we do not aim to improve the classification performance of the $\mathcal{A}_{i}$ model which uses the natural descriptions for inference. Instead, we focus on a) exploring the feasibility of the $\mathcal{A}_{o}$ model which  leverages a different modality to detect Dementia, and b) better understanding which information units from the input space are leaked in the output space contributing to dementia classification. Using these insights, we hope that our work inspires further work on protecting affected populations.


\ignore{

Our threat model considers an adversary $\mathcal{A}$ with black-box access to T2I models. We opted for a black-box approach because it provides a more realistic strategy than a white-box approach, which would necessitate access to the internal components of the T2I model. More specifically, we employ $2$ adversary models with different levels of access to resources, text--based ($\mathcal{A}_{i}$) which uses the natural descriptions as prompts, image--based ($\mathcal{A}_{o}$) which leverages the output images of T2I models. }

\section{Analysis Methods}
\label{sec:evaluation_setup}

\subsection{Overview}
Our analysis comprises three stages: extracting speech-to-text transcriptions; generating images based on those transcriptions; analyzing the relationship between the ability to detect dementia from T2I language inputs and generated images.

\vspace{3pt}\noindent\textbf{Transcriptions.}
We use
ADReSS~\cite{luz2020alzheimer}, a subset of the DementiaBank dataset~\cite{becker1994natural}.
$156$ participants are provided with the same `Cookie Theft Picture' and are asked to describe it orally. The descriptions are manually transcribed eliminating the effect of ASR errors. The ADReSS dataset contains healthy-control (CC) and dementia--labelled (AD) descriptions, with $54$ train and $24$ test samples for each class, for a total of $156$. In our evaluations, 20\% of the train set is set aside as the validation set. All the evaluations were done on the provided test set.


\vspace{3pt}\noindent\textbf{Image Generation.}
We leverage  Stable Diffusion v2.1 an open source T2I model to generate images. Specifically, each sample (description) in the ADReSS dataset is used as an input to the model. Stable Diffusion v2.1 clips the input to 77 tokens which means parts of the descriptions are not considered in the generation. However, this does not affect our analysis which focuses on the difference between AD and CC using the same number of available tokens as the frequency distribution and total count of nouns within the first 77 tokens are nearly identical for both groups. We therefore conclude that restricting the study to the first 77 tokens does not introduce any significant bias.

\vspace{3pt}\noindent\textbf{Analysis.}
The last part is the core of the framework.
For the analysis we (a) develop binary dementia classification models based on text, and based on images; (b) use explainable AI techniques to study what the classifiers learn; (c) identify and introduce new metrics to evaluate the relationship between the input descriptions and the generated images. We describe this in more detail below.

\subsection{Dementia Classification.}
For our classification task, we use all the classic machine learning algorithms integrated in the scikit--learn library~\cite{scikit-learn}. Due to space limitations, we report only our best classifier results in the Evaluation section.
Notably, our classifiers are trained based on embeddings which encode joint representations of text and images in a latent space. This is paramount for our analysis to better identify what characteristics of text are present in the generated images. 
 CLIP~\cite{CLIP} and ViT~\cite{ViT} embeddings are suitable for this purpose. For our classifiers, we use CLIP embeddings.



\subsection{XAI Methods}
Explainable AI (XAI) methods provide insights into the decision-making processes of deep learning models by highlighting the most influential features that drive predictions. For instance, a well--established approach is GradCAM~\cite{GradCAM}, which generates heatmaps to visualize the regions of an input image that contribute most to a model’s prediction. 
Thus, we leverage XAI techniques to understand the contribution of relevant features in the input and output spaces. In this work, we focus our analysis on \textit{Information Units (IU)}. The set of IUs we used is constructed by performing POS tagging on the entire dataset and then collecting the unique and meaningful nouns and verbs.
More specifically, for the input space, we compute the SHAP values for the IUs. For the output space, we use GradCAM on top of SVM to identify the attention score by the classifier over the image. Then, we use Grounded-SAM~\cite{ren2024grounded} to get the mask for the IUs. We compute the contribution score for an IU as the sum of the attention for the mask of the IU divided by the area of the mask. We divide the sum of attention score by the area of the mask so as to normalise any bias towards the contribution score due to the area occupied by the IU. 




\ignore{By applying GradCAM to images generated from dementia-affected speech prompts, we analyze whether T2I models encode and propagate cognitive decline markers. This interpretability framework allows us to assess the extent to which generated images inadvertently reveal sensitive neurocognitive information, reinforcing the need for privacy-preserving generative models.}


\subsection{Metrics}
\label{subsec:evaluation_metrics}
We aim to take a step towards understanding dementia-related information leakage from speech transcriptions to generated images. To this end we evaluate our text--image embedding dementia classifiers in terms of \textit{privacy} and \textit{utility}

\vspace{3pt}\noindent\textbf{Privacy.} We measure dementia leakage via dementia classification accuracy. In other words we assume that if dementia classification is possible in the input or the output space then dementia-related linguistic markers are learned by the classifiers and can be used to infer one's condition. 


\vspace{3pt}\noindent\textbf{Utility.} CLIPScore~\cite{pmlr-v139-ramesh21a} and TIFA~\cite{10377168} are used for measuring the semantic similarity between the input prompts and the output images. For comparing the quality of the images between the two groups, we leverage the Inception Score (IS), FID, and LPIPS. We consider the original cookie theft image as the ground truth image. Furthermore, we introduce two new metrics, namely Information Units Propagation Score (IPS)~\cite{Kumar2023SuperResolutionMetrics} (in Equation~\ref{equation:IPS}) and Extraneous Content Score (ECS) (in Equation~\ref{equation:ecs}) which specifically evaluate the extent to which the desired content in terms of IUs is propagated into the output space and how much of the unintended content appears in the output space respectively. More formally:

\begin{equation}
\label{equation:IPS}
\text{IPS}_{\text{avg}} = \frac{1}{N} \sum_{i=1}^{N} \frac{\sum_{j=1}^{M} \mathbb{1}(\text{IU}_j \in \text{prompt}_i) \cdot \mathbb{1}(\text{C}(\text{IU}_j) \neq 0)}{\sum_{j=1}^{M} \mathbb{1}(\text{IU}_j \in \text{prompt}_i)}
\end{equation}

\begin{equation}
\label{equation:ecs}
\text{ECS}_{\text{avg}} = \frac{1}{N} \sum_{i=1}^{N} \frac{\sum_{j=1}^{M} \mathbb{1}(\text{IU}_j \notin \text{prompt}_i) \cdot \mathbb{1}(\text{C}(\text{IU}_j) \neq 0)}{M - \sum_{j=1}^{M} \mathbb{1}(\text{IU}_j \in \text{prompt}_i)}
\end{equation}
where \emph{M} is total number of IUs, \emph{N} is the size of the test set and \emph{C} is the function which computes the contribution score. Lastly, we note that the contribution scores computed for the IUs are used for evaluation. 
\section{Evaluation}
\label{sec:Evaluation}

\subsection{Research Questions.}
Our evaluation aims to answer the following overall research questions:
\textbf{RQ1:} Can we detect Dementia from generated images?
\textbf{RQ2:} What regions of an image are responsible for the leakage?
\textbf{RQ3:} What features are leaked during the transformation of a prompt to an image?

\subsection{Dementia Leakage}
\label{subsec:dementia_leakage}

To answer RQ1, for the Input model ($\mathcal{A}_i$) we extract the CLIP embedding from text and we train for $100$ epochs, using the Binary Cross Entropy Loss, a Neural Network with $3$ hidden layers and batch size $16$. We apply a $70:30$ training/test split on the dataset and use $20\%$ of the training set for validation to avoid overfitting to the test set. Our $\mathcal{A}_i$ model achieves $83.33\%$ accuracy.
%


\vspace{3pt}\noindent\textbf{Observations.} The results suggest that T2I models can inadvertently reveal sensitive medical information through their outputs. More importantly and in contrast with prior work~\cite{zhu2021exploring}, our $\mathcal{A}_{o}$ approach does not depend on audio information. It merely requires the image generated by the T2I model which is known to the model providers by default. Such images are also publicly exposed when a user shares them online.

\subsection{Salient Regions}
\label{subsec:salient_regions}

To understand how generated images contribute to dementia classification, we analyze the specific image regions or patches that are most influential in the model’s decision--making (RQ$2$). Our analysis is conducted in two settings: (1) within each group separately ($AD$ vs. $CC$), and (2) across both groups collectively to identify shared and distinct classification patterns. This dual approach allows us to highlight common artifacts or regions that are consistently important for classification, as well as group-specific differences in visual features. Inspired by Zhu et al.~\cite{PictureDescriptionAlignment} we also try to understand the contribution of different information units in the image towards classification. However, we use explainability techniques and fine segmentation models, like Grounded SAM, to extract the fine contribution of all the units, rather than relying on selective search, which identifies regions based on another model’s assumptions.

\begin{figure}[!ht]
    \centering
    \subfloat[Comparison of $AD$ and $CC$ groups in the output space.\label{fig:rq2_1}]{
        \includegraphics[width=0.72\linewidth]{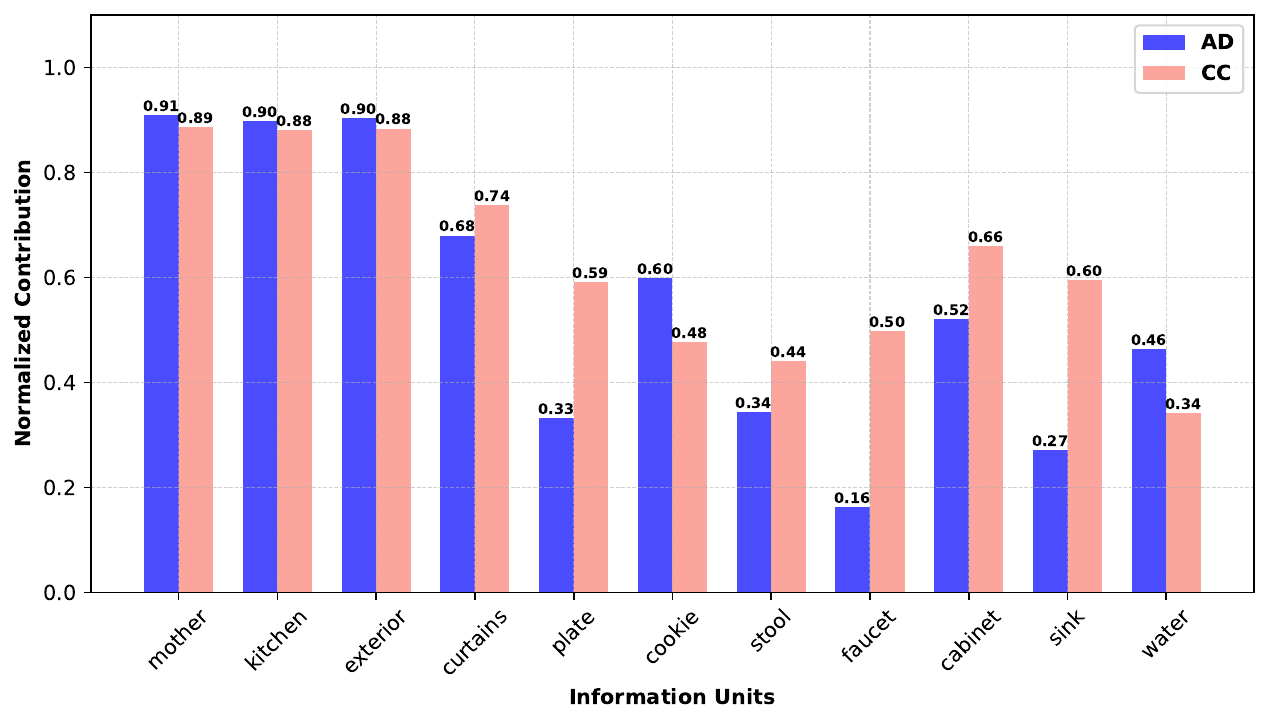}} \vspace{-1em}

    \subfloat[Frequency of the information units in the output space for both groups.\label{fig:rq2_3}]{
        \includegraphics[width=0.72\linewidth]{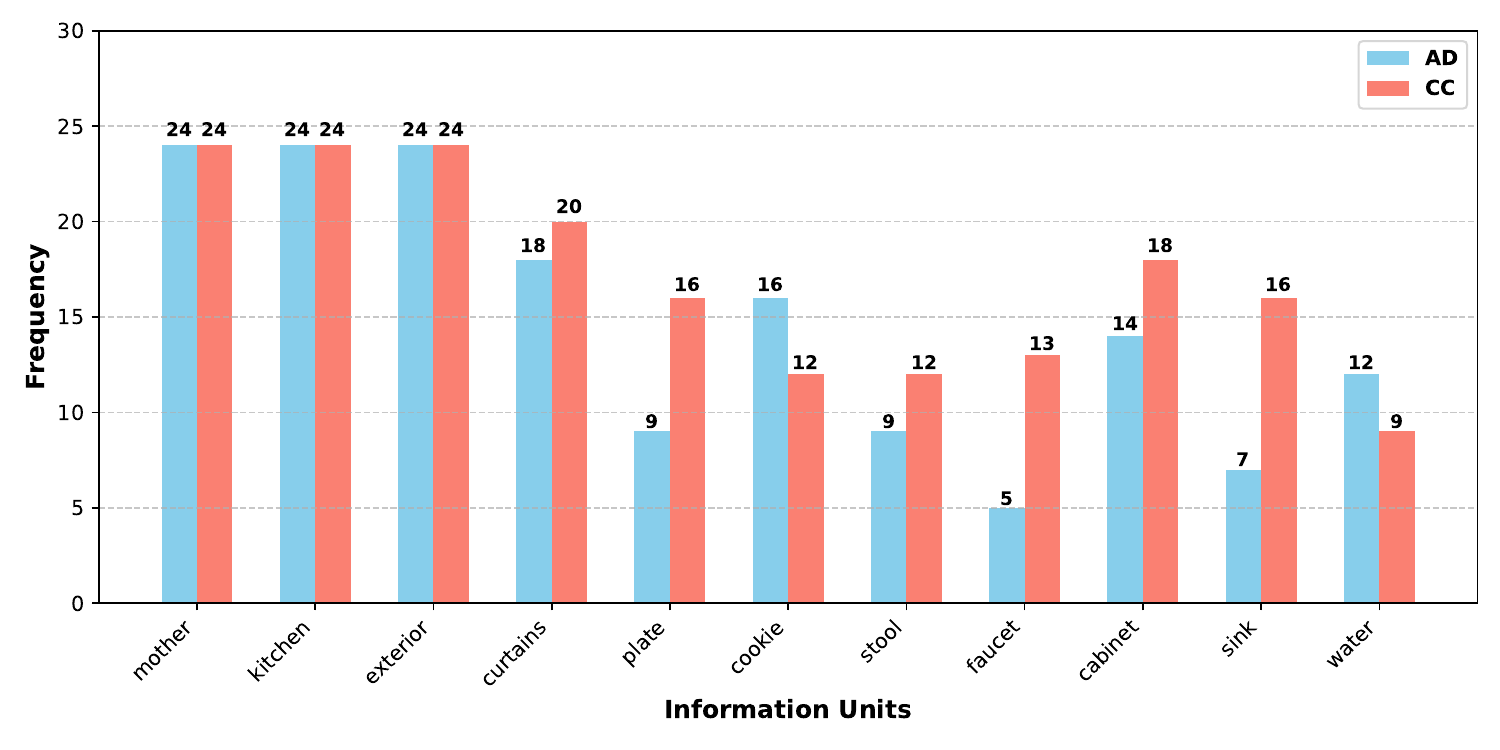}} \vspace{-0.5em}
    \vspace{1em}
    \caption{Analysis of features responsible for classification in the output space.}
    \label{fig:dominant_tokens}
    \vspace{-1.5em}
\end{figure}

Figure~\ref{fig:dominant_tokens} summarizes our results. For space limitation, we include only a few information units which can highlight the trend (in Figure~\ref{fig:rq2_1}). In detail, Figure~\ref{fig:rq2_1} shows that the contribution of IUs like `mother', `kitchen', `exterior' are very high for both classes.\ignore{and those for `stool', `sink',`faucet', and `plate' are low.} More importantly. we see that finer-grained details like `faucet', `stool', `plate', and `sink' are not very important features for the $AD$ group as in $CC$. This same trend is followed by both the groups. Furthermore, Figure~\ref{fig:rq2_3} indicates that the contributions of the features across the two groups directly correspond to the frequency distribution of these information units in the images generated for each group. This suggest that the contribution scores are mostly driven by the representation of specific information units in the output space for both groups.

\subsection{Feature Propagation}
\label{subsec:feature_propagation}
\begin{figure}[!ht]
    \centering
    \subfloat[Comparison of overall contribution of both $AD$ and $CC$ groups in the input v/s output space.\label{fig:rq3_2}]{
        \includegraphics[width=0.72\linewidth]{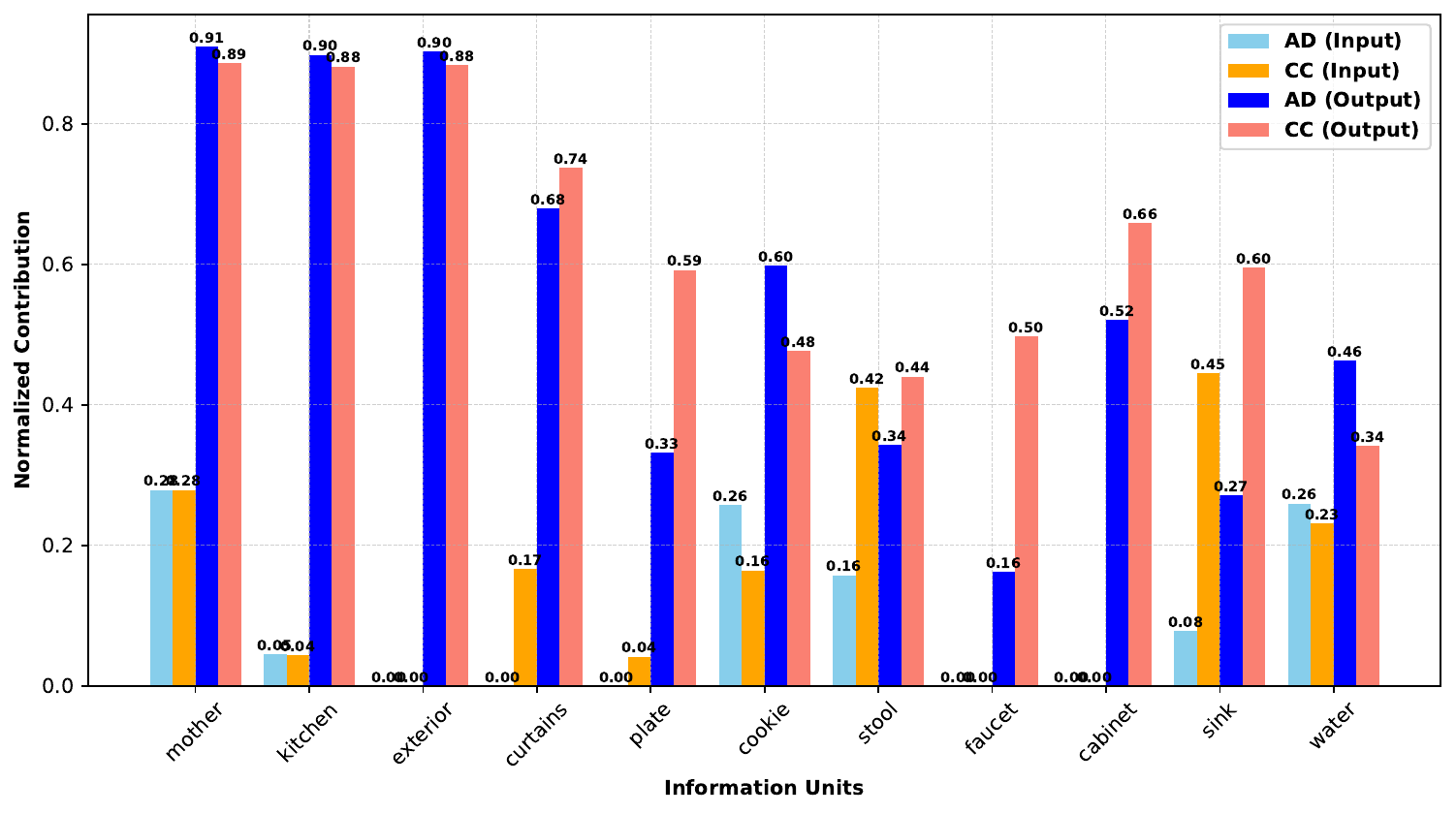}} \vspace{-1em}
    
    \subfloat[Comparison of frequency of information tokens present in the test set for $AD$ and $CC$ groups in the input v/s output space.\label{fig:rq3_1}]{
        \includegraphics[width=0.72\linewidth]{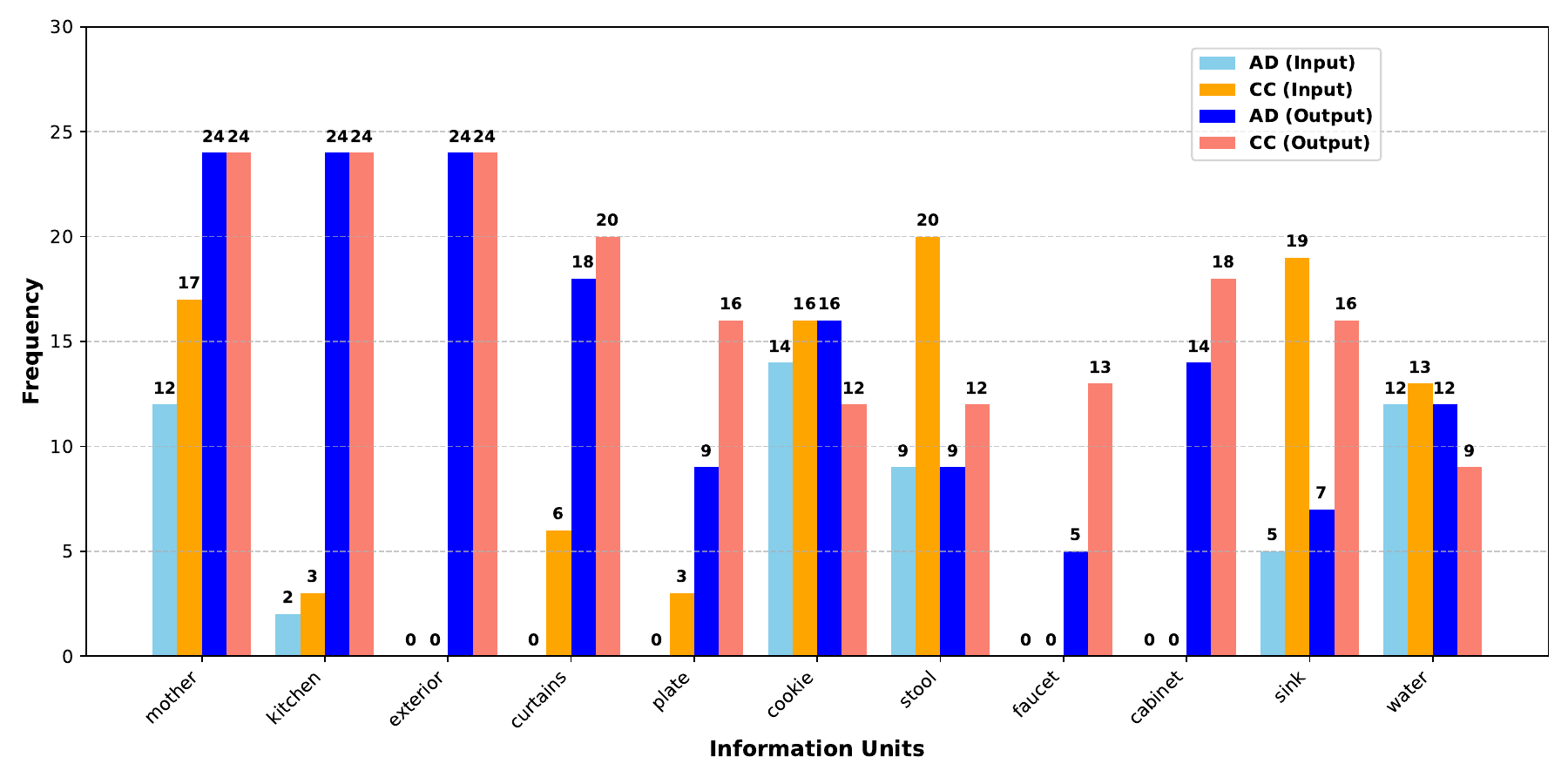}} \vspace{-1em}

    \subfloat[Contribution score of IUs in the output space after removal of discourse tokens.\label{fig:rq3_3}]{
        \includegraphics[width=0.72\linewidth]{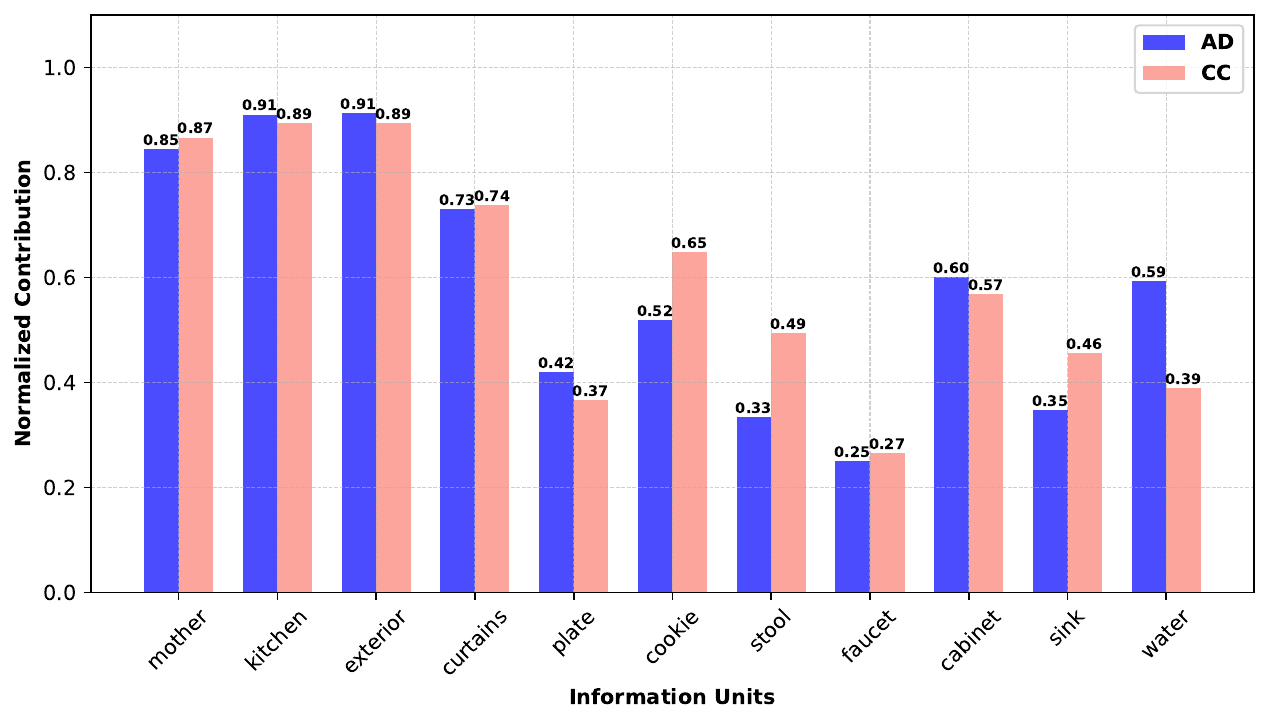}} \vspace{-0.5em}
    \vspace{1em}
    \caption{Comparison of frequency and contribution of IUs across $AD$ and $CC$ in the input as well as output space along with evaluation after removing the discourse tokens.}
    \label{fig:feature_propagation}
    \vspace{-2em}
\end{figure}

To answer RQ3, we focus on analysing how the contribution of IUs to classification changes from the input space to the output space. Strikingly, Figure~\ref{fig:rq3_2} reveals that the most important IUs in the output space were among the least significant in the input space. For instance, the IU `kitchen' which had minimal importance for classification in the input space, emerges as the most critical in the output space. Similarly, units like `faucet' and `cabinet' which were entirely absent in the input space, play a key role in classification within the output space. Beyond that, there is a general increase in the contribution score for all the information units. 

Figure \ref{fig:rq3_1} shows that, similar to the output space, there is a general correspondence between the frequency distribution of information units and their contribution scores in the input space. However, the relationship between the frequency of information tokens in the input space and their actual representation in the output space is highly inconsistent. For example, tokens such as `kitchen', `curtains', `faucet', `cabinet' are scarcely present in the input space but appear prominently in the output space, likely due to the T2I model’s tendency to enhance prompts with additional details. In contrast, information units like `stool', `sink', `water'-—despite being prominent in the input space—-are largely absent in the output space. This suggests that T2I models lose information and introduce noise during the generation process, disrupting the accurate propagation of features into the output space.

We hypothesize that the deviations from the intended generation goal may be influenced by the presence of discourse tokens in the input space. To study their impact, we remove discourse tokens and compute the contribution scores and classification accuracy for the two groups. While the input space accuracy experiences only a slight decline, from $83.33\%$ to $81.25\%$, the output space accuracy drops significantly, from $75\%$ to $62.13\%$. This suggests that \emph{discourse tokens played a crucial role in differentiating the two classes in the output space}. Additionally, we observe that removing the discourse tokens improves the classifier's attention towards minor details in the output bringing the contribution scores for $AD$ and $CC$ closer for most information units (see Figure~\ref{fig:rq3_3}). This reduction in differentiation results in lower classification accuracy in the output space. For example, the contribution score gap of `faucet' between the groups decreased from $0.34$ to $0.02$.

To better understand how the discourse tokens influence the output space, we analyse the fidelity of T2I model outputs across the two groups. Specifically, we examine which elements of the prompts are consistently carried over into the generated images using CLIP, TIFA, and IPS score. The results in Table~\ref{tab:iu_clip_tifa} suggest that removing the discourse tokens improves semantic similarity for both groups. However, interestingly, despite the $CC$ group's prompt being more clear and concise, the $AD$ group exhibits higher semantic similarity in the generated images. This discrepancy can be attributed to significant neglect in T2I diffusion models, where the presence of multiple entities in a prompt increases the likelihood of omitting key elements during generation. In our case, $CC$ prompts contain nearly twice as many entities as $AD$ prompts on average, leading to a greater omission of topics and, consequently, lower semantic similarity.

\begin{table}[!ht]
\small
\vspace{-5pt}
    \scriptsize
    \centering
    \renewcommand{\arraystretch}{1.2}
    \begin{tabular}{c|c|cc|cc|cc}
        \hline
        \multirow{2}{*}{\textbf{Group}} &
      \multirow{2}{*}{\textbf{\begin{tabular}[c]{@{}c@{}}IU \\ Present\end{tabular}}} &
      \multicolumn{2}{c|}{\textbf{CLIP}} &
      \multicolumn{2}{c|}{\textbf{TIFA}} &
      \multicolumn{2}{c}{\textbf{IPS}}\\
         &         & \textbf{Org} & \textbf{ND} & \textbf{Org} & \textbf{ND} & \textbf{Org} & \textbf{ND} \\
        \hline
        $AD$ & $7.20$ & $0.15$ & $0.16$ & $0.28$ & $0.28$ & $0.75$ & $0.77$ \\
        $CC$ & $12.62$ & $0.15$ & $0.15$ & $0.23$ & $0.20$ & $0.76$ & $0.72$\\
        \hline
    \end{tabular}
    \caption{Comparison of IU present, CLIP, TIFA, and IU Propagation Score(IPS) scores for AD and CC groups. \emph{ND} refers to `No Discourse', and \emph{Org} refers to `Original'.}
    \label{tab:iu_clip_tifa}
    \vspace{-1em}
\end{table}

To further analyze the extent of creative freedom exhibited by the T2I model in generating additional content, we compute the ECS score. Table \ref{tab:is_fid_lpips} shows that ECS score decreases from $0.67$ to $0.66$ for the AD group and from $0.75$ to $0.74$ for the CC group, indicating that the removal of discourse tokens from the prompt limited the T2I model's ability to generate extraneous content\footnote{Note that ADReSS dataset has a limited number of data samples, and might not be indicative for large ones.}.

\ignore{
To further analyze the extent of creative freedom exhibited by the T2I model in generating additional content, we compute the ECS score (see Table~\ref{tab:voluntary_addition}). The results suggest that removing discourse tokens reduces the model’s tendency to introduce extra elements into the output.

\begin{table}[!ht]
\vspace{-5pt}
\scriptsize
    \centering
    \renewcommand{\arraystretch}{1.2}
    \begin{tabular}{l|cc}
        \hline
        \multirow{2}{*}{\textbf{Group}} & \multicolumn{2}{c}{\textbf{ECS Score}} \\
        & \textbf{Original} & \textbf{No Discourse} \\
        \hline
        $AD$ & $0.6736$ & $0.6633$ \\
        $CC$ & $0.7505$ & $0.7407$ \\
        \hline
    \end{tabular}
    \caption{Comparison of Extraneous Content Score (ECS) for AD and CC groups before and after removal of discourse tokens.}
    \label{tab:voluntary_addition}
\end{table}
}

Additionally, we evaluate the impact of discourse tokens on generation quality by computing IS, FID, and LPIPS scores for both groups, as summarized in Table~\ref{tab:is_fid_lpips}. An increase in these scores indicate that removing discourse tokens enhances generation quality. In conclusion, while discourse tokens play a significant role in classification within the output space, they are not the sole contributing factor. As shown in Figure~\ref{fig:rq3_3}, even though their removal reduces the gap between the contribution scores of information units across the two groups, a noticeable difference remains. This highlights the presence of additional distinguishing features influencing classification.

\begin{table}[!ht]
\centering
\scriptsize
\vspace{-5pt}
\renewcommand{\arraystretch}{1.2}
\resizebox{\columnwidth}{!}{%
\begin{tabular}{c|cc|cc|cc|cc}
    \hline
    \multirow{2}{*}{\textbf{Group}} & 
    \multicolumn{2}{c|}{\textbf{IS}} & 
    \multicolumn{2}{c|}{\textbf{FID}} & 
    \multicolumn{2}{c|}{\textbf{LPIPS}} & 
    \multicolumn{2}{c}{\textbf{ECS}} \\
    & \textbf{Org} & \textbf{ND} & \textbf{Org} & \textbf{ND} & \textbf{Org} & \textbf{ND} & \textbf{Org} & \textbf{ND} \\
    \hline
    \textbf{AD} & 3.16 & 3.62 & 425.29 & 433.06 & 0.71 & 0.72 & 0.67 & 0.66 \\
    \textbf{CC} & 3.08 & 2.80 & 435.96 & 411.14 & 0.72 & 0.73 & 0.75 & 0.74 \\
    \hline
\end{tabular}
}
\caption{Comparison of IS, FID, LPIPS, and ECS scores for AD and CC groups. \emph{Org} refers to ‘Original’, \emph{ND} refers to ‘No Discourse’.}
\label{tab:is_fid_lpips}
\vspace{-1em}
\end{table}

\ignore{
\textcolor{yellow}{Shouldn't we add this when we talk about metrics itself? We start by evaluating the faithfulness of image generation to the input prompts. To quantify this, we adopt the TIFA (Text--to--Image Faithfulness Assessment) score~\cite{tifa}, a question-answering-based metric that measures the alignment between the generated image and its corresponding textual description.}
}

\section{Conclusion}
\label{sec:Conslusion}

Our work highlights a critical and previously overlooked privacy risk associated with T2I models: the potential for generated images to inadvertently reveal sensitive neurocognitive information. Our findings demonstrate that visual artifacts within these images can serve as unintended indicators of dementia, aligning with linguistic patterns present in speech descriptions. Through the use of explainability techniques, we identified specific regions within the generated images that contribute to classification, reinforcing concerns about implicit leakage of medical conditions through generative AI outputs.
These results underscore the ethical and privacy challenges posed by diffusion models. The ability to infer cognitive health status from generated images raises the risk of unauthorized profiling or discrimination, highlighting the urgent need for mitigation strategies. While adversaries could exploit this vulnerability, various techniques exist to obfuscate sensitive linguistic cues before they are transformed into images~\cite{dominika_arxiv}. Future research should focus on developing robust defenses that ensure the responsible deployment of T2I models, preserving both privacy and inclusivity.

\ignore{

... Our experiments show the potential of using generated images from T2I models  to identify the presence of dementia.

In this work, we investigated the potential privacy risks posed by text-to-image (T2I) diffusion models when processing speech descriptions from individuals with dementia. By leveraging explainability methods such as GradCAM, we demonstrated that certain visual patterns in the generated images correlate with dementia-related linguistic features, raising critical concerns about unintended privacy leakage.

The ability to infer sensitive neurocognitive information from T2I outputs poses ethical and security risks, as these images could be exploited for unauthorized profiling or discrimination. 

This  can be used by attackers but there are different ways to try to obfuscate the text~\cite{dominika_arxiv}.
}


\bibliographystyle{IEEEtran}
\bibliography{bibliography}

\end{document}